\def\year{2020}\relax
\documentclass[letterpaper]{article} 
\usepackage{aaai20}  
\usepackage{times}  
\usepackage{helvet} 
\usepackage{courier}  
\usepackage[hyphens]{url}  
\usepackage{graphicx} 
\usepackage{graphicx}  
\usepackage{booktabs} 
\usepackage{bbm} 
\usepackage{amsmath} 
\usepackage{subfigure} 
\usepackage{color} 

\title{Countering Noisy Labels By Learning From Auxiliary Clean Labels}

\author{%
   Tsung Wei Tsai \\
  \texttt{ccw18@mails.tsinghua.edu.cn} \\
   \And
   Chongxuan Li \\
   \texttt{chongxuanli1991@gmail.com} \\
  \AND
   Jun Zhu \thanks{Correspondence to Jun Zhu: dcszj@mail.tsinghua.edu.cn} \\
   \texttt{dcszj@mail.tsinghua.edu.cn} \\
   \\
   Dept. of Comp. Sci. \& Tech., BNRist Center, \\State Key Lab for Intell. Tech.\& Sys., THBI Lab,
   Tsinghua University, Beijing, 100084, China
}

\begin{document}

\maketitle

\begin{abstract}


We consider the {\it learning from noisy labels} (NL) problem which emerges in many real-world applications. 
In addition to the widely-studied synthetic noise in the NL literature, we also consider the pseudo labels in {\it semi-supervised learning} (Semi-SL) as a special case of NL.
For both types of noise, we argue that the generalization performance of existing methods is highly coupled with the quality of noisy labels. 
Therefore, we counter the problem from a novel and unified perspective: learning from the auxiliary clean labels. 
Specifically, we propose the {\it Rotational-Decoupling  Consistency Regularization} (RDCR) framework that integrates the consistency-based methods with the self-supervised rotation task to learn noise-tolerant representations. The experiments show that RDCR achieves comparable or superior performance than the state-of-the-art methods under small noise, while outperforms the existing methods significantly when there is large noise.

\end{abstract}

\section{Introduction}

Deep neural networks (DNNs) have achieved considerable improvements in learning tasks with voluminous labeled data~\cite{he2016deep}. Nevertheless, collecting high-quality labels is both expensive and time-consuming. The hungriness for labels substantially restricts the prevalence of DNNs in many real-world applications where extensive data comes along with scant labels.

In general, there are two approaches to collect enormous amount of labels, which result in two types of noise in \emph{learning from noisy labels} (NL) literature~\cite{frenay2013classification}. 
Firstly, crowdsourcing provides an inexpensive and efficient way to distribute tedious annotation tasks that require human intelligence to a pool of paid workers~\cite{crowston2012amazon}.
But, the easily accessible label
sets are often corrupted due to malicious or careless workers.
Such noise can be modeled by \emph{noisy at random} (NAR)~\cite{frenay2013classification}, which are widely studied in the form of the synthetic symmetric and asymmetric noise~\cite{han2018co,chen2019understanding}. For clarity, we refer the "simplified-NL" setting as the one involving the synthetic noise.
Secondly, it is a common practice in \emph{semi-supervised learning} (Semi-SL) to assign pseudo labels to the unlabeled data based on predictions of networks pre-trained on limited labeled data~\cite{yarowsky1995unsupervised,lee2013pseudo}.
The pseudo labels are inevitably noisy due to the natural consequence of the generalization error, belonging to the family of \emph{noisy not at random} (NNAR)~\cite{frenay2013classification}. 
Although the pseudo labels in Semi-SL and the synthetic noise in simplified-NL are often studied independently, they are closely related from the view of noisy labels~\cite{frenay2013classification}; therefore, we aim to combat the two types of noise a unified manner.









Even with the recent advances in NL~\cite{chen2019meta,arazo2019unsupervised,lee2019robust},  
the generalization performance is still highly coupled with the quality of noisy labels.
As the quantity of wrong labels increases, the model could hardly distinguish hard examples from the mislabeled ones whilst the mistakes could even reinforce themselves.
According to the experiment results from the literature~\cite{hataya2019unifying,athiwaratkun2018improving,han2018co}, \emph{we notice that having a small subset of clean labels together with vast unlabeled data is superior to having a large amount of noisy labels}, given that the size of data with the true labels is the same.
It clearly evidences that having an identifiable set of clean data points provides a significant performance boost. 

Based on the observation, in this paper, we propose to counter the two types of noise from a novel perspective that has yet been explored: \emph{learning from an auxiliary set of labels which emulates the clean labels by leveraging self-supervised learning (Self-SL)}. This does not require making extra assumptions on the noise. Regardless of the noise level, Self-SL guarantees a certain amount of supervisions from all the inputs. By training on the auxiliary labels simultaneously, we can decouple the noisy labels and further enforce noise-tolerant feature representations.







Specifically, we propose Rotational-Decoupling  Consistency Regularization (RDCR), a unified multi-task framework that integrates the rotation task~\cite{gidaris2018unsupervised} with the consistency-based methods~\cite{tarvainen2017mean,athiwaratkun2018improving,damodaran2019pushing}. 
The consistency-based methods have recently achieved the state-of-the-art results under the simplified-NL and the Semi-SL setting, which improves the robustness to noise by enforcing the flatter loss landscape~\cite{laine2016temporal} and cleansing the noisy training labels~\cite{damodaran2019pushing,chen2019meta}.
We explore surrogate supervisions by augmenting the label set with the easily accessible rotation degrees.
The rotation labels serve as a strong noise regularizer preventing the network from overfitting the noise. 
Our end-to-end formulation also outperforms its pre-trained counterpart~\cite{gidaris2018unsupervised}. 
To encourage more noise-tolerant feature representations, we further investigate and incorporate the group normalization (GN)~\cite{wu2018group} and the weight standardization (WS)~\cite{qiao2019weight}.
We conduct extensive experiments
to demonstrate the robustness of our RDCR on both the synthetic noise and the pseudo labels.

Overall, our contribution is threefold:
(1) We provide a novel perspective that leverages auxiliary clean labels to counter the two types of noise in NL, including the symmetric and asymmetric noise in simplified-NL and pseudo labels in Semi-SL.
(2) We show that we can exploit additional training signals from all the provided images and strengthen the data cleansing mechanism with the self-supervised rotation task.
(3) The proposed RDCR achieves comparable or superior results than the state-of-the-art methods on two types of noise under different noise levels, while outperforms the methods significantly when there is large noise.

\section{Label Noise Models}


The section presents the basic notations and the two categorizations of the noise.
We consider the $K$-class image classification problem of $N$ examples in the presence of label noise. 
We denote the image, true label, observed label and the pseudo label of the $n$-th example respectively as $x_n$, $y_n$, $\widetilde{y}_n$ and $y'_n$, where  $y_n$, $\widetilde{y}_n$ \text{ and } $y'_n \in [K] := \{1, 2, ..., K \}, \forall n \in [N]$. The training set of $N$ examples is denoted by $\mathcal{D} = \{ (x_n, \widetilde{y}_n) \}_{n=1}^N$. 
Under the Semi-SL setting, we have the prior knowledge to identify a small set of $L$ examples with ground truth labels from $\mathcal{D}$, denoted as $\mathcal{D}_L  = \{ (x_l, y_l)\}_{l=1}^L$. The other $N-L$ labels can be regarded as absent or useless. 

We adopt a convolutional neural network (CNN) as the classifier. The mapping of CNN is denoted by $f_{\theta_C}$, followed by a fully connected layer $f_{\theta_S}$ to the space of $K$ categories, where the parameters are shown in the subscript of the functions.  For simplicity, we use $z_n := f_{\theta_C}(x_n)$ to denote the CNN representation of the $n$-th sample. 


Based on~\cite{frenay2013classification},  we categorize label noise into two types
according to the relationship among random variables. We denote $x, y$, and $\widetilde{y}$ as the random variables of the image, the true label, and the observed label respectively. 
The variable $E = \mathbbm{1}(y = \widetilde{y}) $ indicates whether the observed label is corrupted. For interested reader, please refer to~\cite{frenay2013classification} for details. 
The two types of noise are:
\begin{enumerate}
\item {\it Noisy at Random} (NAR): In NAR, the variable $E$ is depending on the true label $y$ while independent to the image. In certain cases, some of the classes are confusingly similar and often lead to asymmetric noise. For simplicity, we will consider the pairwise asymmetric noise~\cite{han2018co}, \textit{i.e.}, $\forall k_1 \in [K]$, we select a class $k_2 \neq k_1$ and randomly annotate some images in class $k_1$ as $k_2$. 

The widely studied symmetric noise~\cite{han2018co,chen2019understanding} can be regarded as a special case of NAR where $E$ is independent of the true label $y$ and the underlying image $x$. 
The generation process of symmetric noise is as follows: A biased coin is flipped to decide whether to change the observed label, in the meanwhile, a dice of $K-1$ dimensions is thrown to assigned the wrong label~\cite{frenay2013classification}. It depicts the situations that the spammers on crowdsourcing platforms intentionally assign random labels.
\item {\it Noisy Not at random} (NNAR): NNAR is the most general and complicated noise type that some of the classes and images are prone to be misclassified. The labeling error is now depending on the underlying image, and the images from different classes that are visually similar can be mistaken with each other. It simulates the real-world label noise such as the one in Clothing1M~\cite{xiao2015learning}. Another less obvious case would be the pseudo labels in Semi-SL. It encodes the knowledge learned by the model; however, the samples in the low-density regions or around the decision boundaries are subject to label error. In this paper, we especially focus on the noise in pseudo labels of consistency-based methods.
\end{enumerate}


For the NAR, the noise can be characterized by a noise transition matrix $T \in \mathbbm{R}^{K \times K}$, where the ($i, j$)-th entry denotes the probability of $i$-th class sample being labeled as class $j$. 
In this paper, We adopt two noise transition matrices that correspond to symmetric and asymmetric noise following~\cite{han2018co,chen2019understanding}.


\begin{figure}
\begin{center}
\includegraphics[width=1.0\columnwidth]{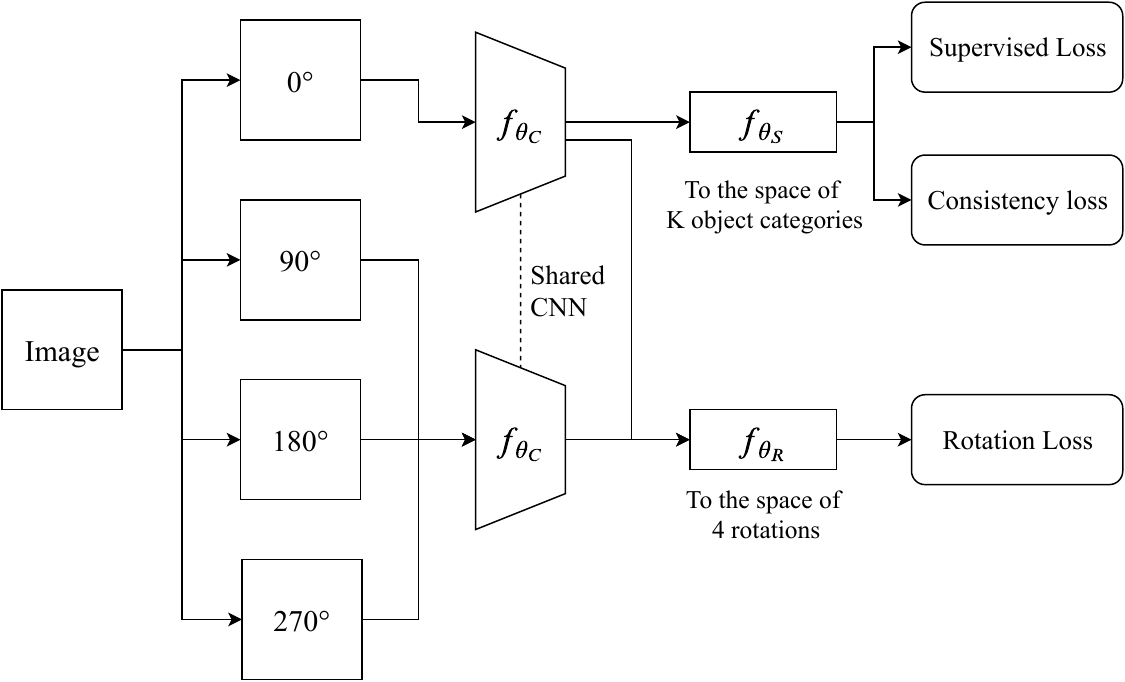}
\caption{Illustration of the RDCR: 
The model receives 4 copies of the same image in different orientations and passes them into the shared convolutional layer $f_{\theta_C}$. The double-head architecture corresponds to 2 sets of predictions, the 1-out-of-K categories, and the four rotation labels. Only the un-rotated images are used to calculate the supervised and unsupervised loss, whilst the rotation loss depends on all the inputs of four orientations.
}
\label{fig:model}
\end{center}
\end{figure}

\section{Rotational-Decoupling Consistency Regularization (RDCR)}




Before presenting the framework, we would like to emphasize the three motivations for our Rotational-Decoupling Consistency Regularization (RDCR).
First of all, to combat the pseudo labels and synthetic noise with a unified approach, 
we propose a generic formulation of the consistency-based methods that encompass some recent works in both NL and Semi-SL literature~\cite{athiwaratkun2018improving,tarvainen2017mean,chen2019meta,damodaran2019pushing}. We select the consistency-based methods as the backbone of RDCR since they improve the robustness to the noise by smoothing the loss landscape~\cite{laine2016temporal,athiwaratkun2018improving} and correcting the noisy training labels~\cite{chen2019meta}.
Next, since the generalization performance of these methods is still heavily relying on noisy labels, we introduce auxiliary clean rotation targets to avoid overfitting the noise while exploiting additional training signal. 
Thirdly, to encourage more noise-tolerant feature representations, we investigate the normalization methods and select an effective combination.





In the following, we begin with the consistency-based method and subsequently describe the proposed RDCR framework together with the normalization methods.

\subsection{Consistency-based methods}

Consistency-based methods have achieved state-of-the-art results under simplified-NL~\cite{chen2019meta,damodaran2019pushing} and Semi-SL setting~\cite{oliver2018realistic,athiwaratkun2018improving}; however, they are often studied independently.
We provide a unified formulation of these approaches. The objective function involves the two loss terms:
\begin{equation}\label{eq1}
    \frac{\omega_S}{S}  \sum_{s=1}^S 
    \mathcal{L}_S( f_{\theta_S}( z_s ) , \widetilde{y}_s ) 
    + 
    \frac{\omega_C}{N} \sum_{n=1}^N\mathcal{L}_C( f_{\theta_S}(z_n), y_n'),
\end{equation}
where the weights $\omega_S$ and $\omega_C$ determine the strength of the two losses, and the latter one often follows the cosine ramp-up schedule~\cite{tarvainen2017mean,laine2016temporal}.
The second term is the unsupervised/consistency loss $\mathcal{L}_C$ that can be either Kullback–Leibler (KL) divergence~\cite{miyato2018virtual} or Mean Square Error (MSE)~\cite{laine2016temporal}. 
The first term is the supervised loss $\mathcal{L}_S$ that computes the cross-entropy between predictions and the observed labels on the selected subset $\mathcal{D}_S \subset \mathcal{D}$ of size $S$.  
For the simplified-NL and the Semi-SL settings, we use $\mathcal{D}$ and $\mathcal{D}_L$ respectively as the selected subset $\mathcal{D}_S$~\cite{athiwaratkun2018improving,chen2019meta}. 
Regarding the type of noise one may encounter, NNAR is unavoidable in both settings due to the consistency term, while the former one may face an additional NAR from the noisy dataset $\mathcal{D}$.



The class of methods deals with the noisy supervisions by encouraging the flatter loss landscape and cleansing the noisy training data.
Predictions under different model-space and input-space perturbations are required to be consistent~\cite{bachman2014learning}. 
Each of the input is at least evaluated twice, which results in a student and a teacher prediction for computing the consistency loss.
The relatively stable teacher predictions serve as pseudo/target labels for the student predictions, which are often cleaner than the observed labels.
\subsection{Decoupling with rotation labels}

Even though the existing consistency-based methods demonstrate certain robustness with respect to label corruptions, we still observe significant performance degradation under the extreme noise because of the heavy reliance on noisy labels. 
Enlightened by the significance of clean labels, we extend the previous framework with an auxiliary self-supervised rotation task to extract additional supervisions from all the provided images and to strengthen the data cleansing mechanism. 


By applying four rotation transformations to the $N$ examples, we can obtain a set of transformed images $\mathcal{D}_R = \{ (x_{r}^R, y_{r}^R) \}_{r = 1}^{4 N}$ following~\cite{gidaris2018unsupervised}. 
The augmented self-supervised label $y^R_r \in [4],  \forall r = 1, ..., 4N$, and the four classes correspond to the 0, 90, 180 and 270 rotation degrees respectively.

In order to achieve higher performance than models trained separately on each task~\cite{athiwaratkun2018improving,gidaris2018unsupervised}, we adopt the multi-task learning framework. The two tasks correspond to making predictions in the space of $K$ object categories and in the space of the four orientations. In this way, the optimization objective becomes the combination of the supervised loss on $\mathcal{D}_S$, the consistency loss on $\mathcal{D}$, and the rotation loss on $\mathcal{D}_R$, where each of them incurs gradients flow back to the joint CNN structure. 
The joint training scheme encourages more robust representations across tasks, at the same time, calibrating discriminative representations for each specific task~\cite{liu2018end}. The objective function of the RDCR extends the Eq.~\ref{eq1} with the rotation loss $\mathcal{L}_R$:
\begin{equation}
\begin{split}
    \frac{\omega_S}{S}  \sum_{s=1}^S 
    \mathcal{L}_S( f_{\theta_S}( z_s ) , \widetilde{y}_s ) 
    + 
    \frac{\omega_C}{N} \sum_{n=1}^N\mathcal{L}_C( f_{\theta_S}(z_n), y_n')
    +\\
    \frac{\omega_R}{4N} 
    \sum_{r=1}^{4 N} 
    \mathcal{L}_R( f_{\theta_R}( z_r ) , y_r^R ), 
\end{split}
\end{equation}
%
where $f_{\theta_R}$ is an extra fully connected layer that maps the hidden features to the rotation targets and $\omega_R$ is the weight of the rotation loss $\mathcal{L}_R$. Following~\cite{gidaris2018unsupervised}, we use the cross-entropy function for $\mathcal{L}_R$.

We attribute the superiority of our method to two key properties of the self-supervised rotation labels, which supplements capabilities of the supervised and consistency loss.

Firstly, we are able to accumulate a considerable amount of supervisory signals by leveraging the vast unlabeled or noisily-labeled images, notwithstanding the noise level.
In essence, the rotation loss is forcing the model to be \textit{rotation covariance}, \textit{i.e.}, given rotated images, the model produces the corresponding labels according to the predefined mapping between the angles and the labels~\cite{marcos2017rotation}.
In order to identify the orientation, the model is required to perceive the objects appear in the images.

Secondly, the rotation labels reinforce the data cleansing mechanism by enhancing the quality of pseudo labels. Without making extra assumptions on the noise, the noisy labels are assessed and benchmarked against the rotation labels.
One may argue that the rotation labels are also corrupted as some objects are rotation agnostic~\cite{feng2019self}, but the noise is actually negligible under the multi-task framework. We specifically conduct an experiment to verify our claim in Sec.~\ref{exp:pseudo}.

Overall, we can decouple the noise presented in the observed and pseudo labels in a unified manner, while including the rotation loss provides strong regularization against incorrectly labeled data.





\subsection{Inducing Noise-tolerant Representations}

We adopt two normalization methods, the weight standardization (WS)~\cite{qiao2019weight} and the group normalization (GN)~\cite{wu2018group}, to induce more robust representations, despite they are originally proposed neither for rotated inputs nor the noisy labels.  



On one hand, WS is introduced to supplement the consistency loss as it can further smooth the loss landscape via decreasing the Lipschitz constants of the loss~\cite{qiao2019weight}.
Considering that the incorrect labels tend to distort and sharpen the decision boundary, WS can help to enforce a more consistent local neighborhoods and thus merge images into coherent clusters~\cite{laine2016temporal}.



On the other hand, given that the batch size is quadrupled with the rotation transformations, we conjecture that it is unnecessary or even harmful to perform normalization on all the rotated inputs using the batch normalization (BN)~\cite{ioffe2015batch}. 
GN bypasses the issue through computing the normalization statistics within each group of channels, which is independent of the batch sizes.
We empirically justify that GN generalizes better than BN when the rotation loss is included. 

With GN and WS, we improve the robustness and generalization performance with few computational overheads. The full comparisons among the normalization methods are in Sec.~\ref{ablation:NL}.



\begin{table}[]
\caption{Test accuracy (\%) on CIFAR-10 (top) and CIFAR-100 (bottom). The "Best" results show the highest value achieved during training, the "Last" results are the performances at the end, and the "CE" represents the vanilla training merely on observed labels with the cross-entropy loss. The experiments consider the NAR in the form of symmetric and asymmetric noise.}
\centering
\label{tab:NL}
\resizebox{1.0\columnwidth}!{
\begin{tabular}{ll|lll|l}
\toprule
Model &      & \multicolumn{3}{c|}{Sym.} & \multicolumn{1}{c}{Asym.} \\
                               &          & 20\%      & 50\%      & 80\%     & 40\%         \\ \hline
CE                       & Best  & 91.49    & 86.62   & 58.19   & 89.95      \\
                               & Last  & 84.04    & 62.45    & 25.15   & 59.74      \\ \hline
MT                             & Best  & 93.55    & 90.08    & 61.11   & 90.80      \\
                               & Last & 92.96    & 76.66    & 24.81   & 77.81      \\ \hline
RD-MT                         & Best & 94.17    & 93.00    & 84.00   & 91.74      \\
(Ours)                               & Last & \textbf{94.17}     & 92.85   & 84.00   & 90.97      \\ \hline
RD-fast-SWA                        & Best     &  \textbf{94.34}   & \textbf{93.17}   & \textbf{84.47} & \textbf{92.68}  \\
(Ours)                               & Last       & 94.14    & \textbf{92.94}    & \textbf{84.23}   & \textbf{92.64}      \\ \bottomrule
\end{tabular}}
\resizebox{1.0\columnwidth}!{
\begin{tabular}{ll|lll|l}
\toprule
Model  &       & \multicolumn{3}{c|}{Sym.} & \multicolumn{1}{c}{Asym.} \\
                               &         & 20\%      & 50\%      & 80\%     & 40\%         \\ \hline
CE                       & Best  & 69.34    & 61.28    & 34.81   & 58.29      \\
                               & Last & 63.14    & 42.45    & 18.67   & 46.41      \\ \hline
MT                             & Best  & 70.46    & 61.87    & 35.12   & 50.26      \\
                               & Last  & 63.35    & 43.60    & 18.55   & 58.75      \\ \hline
RD-MT                         & Best  & 70.93    & 67.21    & 53.99   & 59.60      \\
(Ours)                              & Last  & 70.84    & 66.75    & 53.86   & 59.59      \\ \hline
RD-fast-SWA                           & Best  & \textbf{73.66}    & \textbf{68.21} & \textbf{56.06}   & \textbf{63.40}      \\
(Ours)                               & Last  & \textbf{73.66}    & \textbf{67.33}  & \textbf{56.04}   & \textbf{63.40}      \\ \bottomrule
\end{tabular}}
\end{table}

\section{Experiments}

We compare our RDCR with the state-of-the-art models in both NL and Semi-SL literature on CIFAR-10 and CIFAR-100 datasets~\cite{krizhevsky2009learning}.
Both benchmarks consist of 32-by-32 RGB natural images. 
CIFAR-10 and CIFAR-100 are composed of 50,000 training images and 10,000 testing images from 10 and 100 classes respectively. 
We adopt the MT~\cite{tarvainen2017mean} and fast-SWA~\cite{athiwaratkun2018improving} models for the consistency regularization part of our implementations, which leads to two instances of the RDCR framework: {\it Rotational-Decoupling-MT} (RD-MT) and {\it Rotational-Decoupling-fast-SWA} (RD-fast-SWA). 
All the MT-based models are trained for 180 epochs. We use the 13-layer CNN following the common practice~\cite{tarvainen2017mean,athiwaratkun2018improving} for fair comparisons. Except that the BN layers are replaced by the GN+WS ones in which the number of channels is set to 16 as suggested by~\cite{wu2018group}.
Unless specified, the hyper-parameters of the consistency regularization part exactly follow those of the original MT or fast-SWA implementations. We will leave the specific configuration for each setting in the subsections. 

For all the experiments, we retain 1$\%$ of the training data for validation. It is common and realistic to have a tiny validation set~\cite{ren2018learning,oliver2018realistic}. 

In addition, the complex interaction between the three loss functions invokes the exploding gradient problem described in~\cite{bengio1994learning}. 
We introduce the gradient norm clipping proposed by~\cite{pascanu2013difficulty} to scale down the gradients if the Euclidean norm exceeds the predefined threshold. To be specific, we set $\nabla f = \frac{\tau}{|| \nabla f ||} \nabla f$ for any gradient $\nabla f$ whose norm is greater than the threshold $\tau = 3.0$.

Overall, the experiments show that:

\begin{itemize}
    \item 
    For the synthetic symmetric and asymmetric noise, our methods are superior or comparative to the existing NL methods under all noise levels. We attain significant improvements for noise level greater than 60\%.
    \item As RDCR deals with the pseudo labels, it is applicable to Semi-SL. In the Semi-SL setting, our methods consistently outperform baseline methods, especially when fewer clean labels are presented.
\end{itemize}

\begin{table}[]
\centering
\caption{Test accuracy (\%) on CIFAR-10 (top) and CIFAR-100 (bottom) under four levels of symmetric noise. Note that each method may use different architectures and sizes of the validation set.
}
\label{tab:NLbase}
\resizebox{1.0\columnwidth}!{
\begin{tabular}{@{}llllll@{}}
\toprule
Model         &    0\% & 20\% & 40\% & 60\% & 80\% \\ \midrule
MentorNet~\cite{jiang2017mentornet}  &     & 92.0 & 89.0 &      & 49.0 \\
D2L~\cite{ma2018dimensionality}     & 89.4  & 85.1 & 83.4 & 72.8 &      \\
Reweighting~\cite{ren2018learning}    &    &      & 86.9 &      &      \\
Iterative~\cite{wang2018iterative}   & & 81.4 & 78.2 &      &      \\
WAT~\cite{damodaran2019pushing} & 91.9 & 89.1 & 84.6 & & \\
RoG~\cite{lee2019robust}        & 94.2   & 87.4 & 81.8 & 75.5 &      \\
INCV~\cite{chen2019understanding}         & 89.7 &      &      & 52.3 \\
M-DYR-H~\cite{arazo2019unsupervised}
    & 93.4    & 94.0 & 92.8 & 90.3 & 46.3 \\
MD-DYR-SH~\cite{arazo2019unsupervised} & 92.7 & 93.8 & 92.3 & 86.1 & 74.1 \\
RD-MT (Ours) & 94.4   & \textbf{94.2} & \textbf{93.5} & 88.9 & 84.0 \\
RD-fast-SWA (Ours)          &  \textbf{94.6}          & 94.1  & \textbf{93.5} & \textbf{91.6} & \textbf{84.2} \\ \bottomrule
\end{tabular}}
\resizebox{1.0\columnwidth}!{
\begin{tabular}{@{}lllllll@{}}
\toprule
Model       &    0\%   & 20\% & 40\% & 60\% & 80\% \\ \midrule
MentorNet~\cite{jiang2017mentornet} &  & 73.0 & 68.0 &      & 35.0 \\
D2L~\cite{ma2018dimensionality}     & 68.6 & 62.2 & 52.0 & 42.3 &      \\
Reweighting~\cite{ren2018learning} &   &      & 61.3 &      &      \\
WAT~\cite{damodaran2019pushing}  & 68.2 & 62.7 & 58.9 & & \\
RoG~\cite{lee2019robust}        & 77.0    & 68.3 & 60.8 & 48.4 &      \\
M-DYR-H~\cite{arazo2019unsupervised} & 66.2 & 70.0 &  64.4& 58.1 & 45.5 \\
MD-DYR-SH~\cite{arazo2019unsupervised}  & 71.3 & \textbf{73.7} & \textbf{70.1} & 59.5 & 39.5 \\
RD-MT (Ours)     &  77.1  & 70.8 & 69.4 & 62.3 & 53.9 \\
RD-fast-SWA (Ours)      &  \textbf{77.6}     & \textbf{73.7}  & 69.4 & \textbf{64.3} & \textbf{56.0} \\ \bottomrule
\end{tabular}}
\end{table}

\begin{table*}[t]
\centering
\label{table:cifar10}
\caption{
CIFAR-10/100 test error rates (\%) with a 13-layer CNN under the different number of labels. The number of labels decides the noise level of the pseudo labels, \emph{i.e.}, NNAR.
The first sector presents the results from the literature, the second one includes the baseline models we implement by replacing BN with GN+WS, and the third one displays the self-supervised baselines following~\cite{gidaris2018unsupervised}.
The last sector is the results of the proposed framework.
Noth that the choice of the normalization layer and the number of training epochs are shown in the parenthesis. For the last three sectors, the mean and standard deviation are computed on three runs.
}
\begin{tabular}{lllll}
\toprule
Model / Dataset-$\#$labels        & CIFAR10-1K      & CIFAR10-2K        & CIFAR10-4K    & CIFAR100-10K  \\ 
\midrule
Supervised-only~\cite{tarvainen2017mean}    & 46.43 $\pm$ 1.21 & 33.94 $\pm$ 0.73 & 20.66  $\pm$ 0.57 & 44.56 $\pm$ 0.30\\
$\Pi$ model~\cite{tarvainen2017mean}     & 27.36 $\pm$ 1.20 & 18.02 $\pm$ 0.60 & 13.20 $\pm$ 0.27  & 39.19 $\pm$ 0.36\\
TempEns~\cite{laine2016temporal}        &                &                & 12.16 $\pm$ 0.24  & 38.65 $\pm$ 0.51\\
VAdD~\cite{park2018adversarial}                                &                &                & 9.22 $\pm$ 0.10  \\
VAT-EntMin~\cite{miyato2018virtual}                        &                &                & 10.55           \\
SNTG~\cite{luo2018smooth}                                & 18.41 $\pm$ 0.52 & 13.64 $\pm$ 0.32 & 9.89 $\pm$ 0.34   \\
DCT~\cite{qiao2018deep}                    &                &                & 8.35 $\pm$ 0.06   & 34.63 $\pm$ 0.14\\
Tri-Net~\cite{chen2018tri}            &                &                & \textbf{8.30} $\pm$ \textbf{0.15}   \\
MT (BN, 180)~\cite{athiwaratkun2018improving}     & 18.78 $\pm$ 0.31 & 14.43 $\pm$ 0.20 & 11.41 $\pm$ 0.27  & 35.96 $\pm$ 0.77\\
fast-SWA (BN, 1200)~\cite{athiwaratkun2018improving}      & \textbf{15.58} $\pm$ \textbf{0.12} & \textbf{11.02} $\pm$ \textbf{0.23} & 9.05 $\pm$ 0.21  & \textbf{33.62} $\pm$ \textbf{0.54}  \\ \midrule
MT (GN+WS, 180)                        & 18.08 $\pm$ 1.66 & 13.22 $\pm$ 0.02 & 10.81 $\pm$ 0.09  & 36.62 $\pm$ 0.43\\
fast-SWA (GN+WS, 180)                  & 17.54 $\pm$ 1.64 & 12.83 $\pm$ 0.26 & 10.13 $\pm$ 0.16  & 35.96 $\pm$ 0.58\\
fast-SWA (GN+WS, 780)                  & 16.66 $\pm$ 1.20 & 12.12 $\pm$ 0.06 & 9.61 $\pm$ 0.04  & 34.78 $\pm$ 0.43\\ 
\midrule
Rot + Linear (GN+WS, 180)                   &30.79 $\pm$ 0.42  & 29.90 $\pm$ 0.35 & 30.12 $\pm$ 0.32 & 87.78 $\pm$ 0.01  \\
Rot + Fine-tune (GN+WS, 780)    & 12.16 $\pm$ 0.04 &  9.78 $\pm$ 0.36 &  8.45 $\pm$ 0.27 &  34.55 $\pm$ 0.01 \\
\midrule
RD-MT (GN+WS, 180)                  & 9.39 $\pm$ 0.12 & 8.58 $\pm$ 0.36  & 7.50 $\pm$ 0.24  & 32.43 $\pm$ 0.25   \\
RD-fast-SWA (GN+WS, 180)  & 9.12 $\pm$ 0.23 & 8.29 $\pm$ 0.18  & 7.14 $\pm$ 0.24      & 31.80 $\pm$ 0.43          \\
RD-fast-SWA  (GN+WS, 780)  & \textbf{8.70} $\pm$ \textbf{0.09}  &  \textbf{7.88} $\pm$ \textbf{0.14} & \textbf{6.80} $\pm$ \textbf{0.29}     & \textbf{30.62} $\pm$ \textbf{0.20}
\\
\bottomrule
\end{tabular}
\end{table*}

\subsection{Synthetic noise (simplified-NL)}

To verify the effectiveness of our methods, we conduct extensive experiments on the two datasets on symmetric and asymmetric noise under different noise levels. Although the pseudo labels are also involved due to the consistency term, they are not the main issue given that the synthetic noise are noisier. We trained MT and fast-SWA model for 180 and 360 epochs respectively.

To decouple the noisy labels gradually, we use different weight scheduling schemes for each weight. The supervised weight $\omega_S$ follows the cosine ramp-down schedule~\cite{loshchilov2016sgdr} to reduce the effect of observed labels in the later stage. In CIFAR-100, we use a higher $\omega_R$ to rely more on pseudo labels. It is set to 1000 for $40\% \sim 60\%$ symmetric noise and 40\% asymmetric noise, while for 80\% symmetric noise, we set it to 10,000.
Lastly, for noise level smaller or equal to 60\%, the rotation weight $\omega_R$ follows the linear ramp-up schedule starting from 0 to 0.3. For 80\% symmetric noise, $\omega_R$ ramp-up from 0.3 to 0.5.
With the linear ramp-up, the model can gradually focus on the clean rotation labels at the later stage of training, which help to retain the generalizability attained in the early stage from the noisy labels.

Our methods consistently outperform the MT model, as shown in Tab.~\ref{tab:NL}. The higher "Best" values indicate that our methods exploit more valuable information from the data. The gap between "Best" and "Last" can be viewed as a rough quantification of the negative effect of noise on generalization, we obtain smaller gaps than the baselines.

In addition, Tab.~\ref{tab:NLbase} compares with the state-of-the-art methods under four levels of symmetric noise. Note that these methods may use different architectures and different sizes of the clean validation set, so the results should be interpreted carefully. Nevertheless, we can see that when the noise is large, we obtain significantly higher performance. For noise smaller than 60\%, the adverse effect of noise is even unapparent.

\paragraph{Verification of the training data cleansing mechanism.}~\label{exp:pseudo}

To verify that the rotation labels reinforce the cleansing process of the training data, we can examine the quality of pseudo labels used to substitute the observed labels.
We consider a worse case of 80\% symmetric noise. MT model can hardly infer correct labels based on the highly corrupted observed labels, results in erroneous pseudo labels as shown in Fig.2 (a). Note that the confusion matrix is calculated on the pseudo labels of the training data.
If no extra knowledge is given, the mistakes often reinforce each other and prevent the model to correct itself. 
The error in Fig.2 (a) resembles the symmetric noise since MT memorizes most of the observed labels.
In contrast, our method produces a concentrated confusion matrices where most of the pseudo labels are correct, which restores the data cleansing mechanism MT strives for. 
This directly leads to lower generalization error, as shown in Tab.~\ref{tab:NL}.

\begin{figure}{}
\begin{center}
\label{cm}
\subfigure[MT]{\includegraphics[width=0.495\columnwidth]{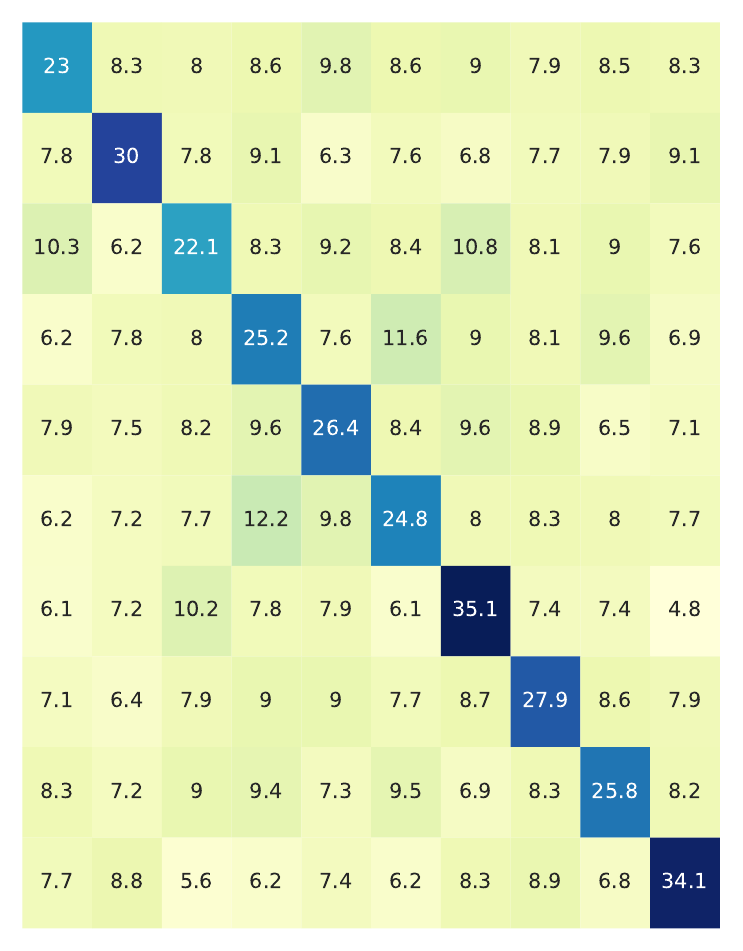}}
\subfigure[RD-MT]{\includegraphics[width=0.495\columnwidth]{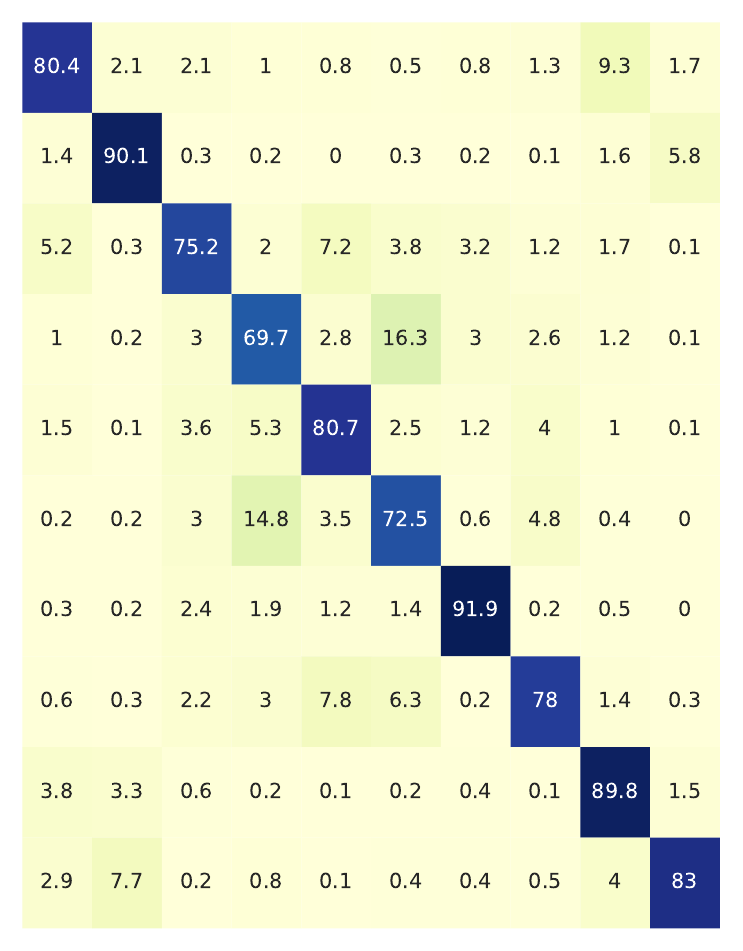}}
\end{center}
\caption{Verification of the data cleansing mechanism by comparing the confusion matrices of pseudo labels on the CIFAR-10 training data between MT model and our RD-MT. 
We use the 80\% symmetric noise here and both models are trained 180 epochs.
The x-axis is the predicted label, the y-axis is the true label, and the values represent the ratio (\%). The higher the ratios in the diagonal indicate the better robustness and noise-correction capability.}
\end{figure}






\subsection{Pseudo Labels (Semi-SL)}

In order to evaluate the robustness of our framework against the pseudo labels, we follow the regular Semi-SL setting which we have vast unlabeled data and a small set of correctly labeled data. 
For consistency-based methods, the pseudo labels are the only type of noise involved. The fewer the clean labels, the higher is the noise in pseudo labels.
For CIFAR-10, we consider the cases where only 1000, 2000, and 4000 clean labels are given; for CIFAR-100, we test the 10,000 labels case. Note that most of the baselines in Tab.3 use a similar 13-layer CNN, we intentionally select the architecture for fairness. 
We also include the self-supervised pre-trained and fine-tuned baselines. For the two baselines, the networks are pre-trained only with the rotation targets for 180 epochs with $\omega_R$ fixed to 10/1 for CIFAR10/100. 
"Rot + Linear" is the pure Self-SL baseline, where we train a fully connected layer on top while keeping the other layers fixed for another 180 epochs. For "Rot + Fine-tune", we follow the training protocol of fast-SWA.

With regard to the hyper-parameters, we apply another 20 learning rate cycles with 30 epochs each based on the pre-trained MT to get the fast-SWA model, \textit{i.e.}, 780 epochs in total.
The weight  $\omega_S$ for the supervised loss is always set to $1$. The $\omega_C$ follows the same cosine ramp-up schedule as the original MT and fast-SWA. As for rotation weight $\omega_R$, it is set constant to $10$ and $1$ in all CIFAR-10 and CIFAR-100 experiments respectively.

The results are displayed in Tab.3. Our method outperforms the state-of-the-art results on the two benchmark datasets under the different number of clean labels. To be specific, for the fast-SWA model, we decrease the error rate from 9.05$\%$ to 6.80$\%$ and from 33.62$\%$ to 30.62$\%$ for CIFAR-10 and CIFAR-100 respectively.

Remarkably, the improvements are even larger when there are fewer labeled samples, \textit{i.e.}, higher pseudo label noise. 
With merely half or one-fourth of the labels, we are able to achieve better performance than the baselines using 4000 labels. The performance of the consistency-based models relies on the number of labels to a great extent. However, for RDCR, the quantity makes relatively little differences as we have access to the informative self-supervised labels for all the data. 




\begin{table}[t]
\centering
\label{table:layer}
\caption{Comparing the test error (\%) of different normalization layers using a 13-layer CNN. We run the experiments run on CIFAR-10 with 4000 labels for 180 training epochs based on the MT model.}
\resizebox{1.0\columnwidth}!{
\begin{tabular}{@{}lccccc@{}}
\toprule
Model & BN    & IN    & LN  & GN & GN+WS (Ours)\\ \midrule
MT                  & 11.48 & 11.60 & 11.47  & 11.27        & \textbf{10.83}    \\
RD-MT            & 9.51      &  9.98     & 8.73    & 8.09 &  \textbf{7.45}           \\ 
\midrule
Differences  & 1.97 & 1.62 & 2.74 & 3.19 & \textbf{3.38} \\
\bottomrule
\end{tabular}}
\end{table}
\paragraph{Ablation Studies.}\label{ablation:NL}
We compare GN+WS against the instance normalization (IN)~\cite{ulyanov2016instance}, the layer normalization (LN)~\cite{lei2016layer}, GN, and the originally used BN, as shown in Tab.4. IN and LN can be viewed as the special cases of GN, where the number of groups is set to one and the number of channels respectively. 

Including the rotation loss consistently improves the accuracy regardless of the underlying normalization layers, while we gain the greatest improvement with GN+WS.



\section{Related Work}
We will briefly discuss the papers in NL, Semi-SL, and Self-SL that are closely related to our model.

{\bf NL:} It remains challenging to learn robustly against the ubiquitous noisy labels as DNNs can memorize the noise with their high capacity~\cite{zhang2016understanding}.
A set of researches focus on designing criteria to select and re-weight samples to avoid over-training on noisy labels, including but not limited to MentorNet~\cite{jiang2017mentornet}, Co-teaching~\cite{han2018co}, and Reweighting~\cite{ren2018learning}. These methods suffer from a waste of samples 
~\cite{chen2019meta}.
In addition to the above methods, \citeauthor{lee2019robust} apply a generative classifier on a pre-trained network for robust inference,~\citeauthor{tanaka2018joint} propose to alternatively update the network parameters and labels, and \citeauthor{arazo2019unsupervised} model the loss with a two-component mixture model.
Another set of researches extend consistency regularization to deal with label noise, where \citeauthor{damodaran2019pushing} introduce a Wasserstein distance, and \citeauthor{chen2019meta} formulate a meta manifold regularizer. 
Similar to our setting, \citeauthor{hataya2019unifying} consider bi-quality data that includes a small set of clean labels and some noisily-labeled data.

{\bf Semi-SL:} There are extensive methods in the  Semi-SL literature~\cite{chapelle2009semi} that improve generalization with the unlabeled data, including but not limited to self-training~\cite{yarowsky1995unsupervised}, generative~\cite{chongxuan2017triple}, and disagreement-based~\cite{zhou2005tri} models. 
The consistency-based~\cite{tarvainen2017mean,luo2018smooth} approaches aim to train classifiers that are robust to random perturbations~\cite{bachman2014learning} by enforcing consistent local neighborhoods. 
TempEns~\cite{laine2016temporal} and MT~\cite{tarvainen2017mean} aggregate the past predictions and weights respectively by EMA 
fast-SWA~\cite{athiwaratkun2018improving} averages selected points traversed along the cyclical learning trajectories with equal weighting.
Recently, the concurrent work~\cite{Zhai2019Smathbf4LSS} also apply Self-SL to consistency-based methods. Their work is developed independently of ours and is limited to only the Semi-SL setting. 
Note that the pseudo labels in self-training~\cite{yarowsky1995unsupervised}, pseudo-labeling~\cite{lee2013pseudo}, and consistency-based methods belong to NNAR.

{\bf Self-SL:} 
The Self-SL researches focus on designing specific unsupervised pre-training objectives that are beneficial to the downstream tasks, \textit{e.g.}, depth prediction, object detection, and image classification~\cite{kolesnikov2019revisiting,doersch2017multi}. For image classification, the learning task can be solving a jigsaw puzzle~\cite{noroozi2016unsupervised}, counting visual primitives~\cite{noroozi2017representation}, and colorizing gray-scale photos~\cite{larsson2017colorization}.
Image rotation task has been applied to image generation~~\cite{lucic2019high} and Semi-SL image classification~\cite{Zhai2019Smathbf4LSS}. Nevertheless, the paradigm has yet to be explored thoroughly in NL. 
\section{Conclusions and Future Work}
In this paper, we propose a unified framework RDCR to deal with the synthetic noise in simplified-NL and the pseudo labels in Semi-SL. Our RDCR decouples the noisy labels to stimulate the data cleansing process and to exploit extra supervisions from all the inputs. Extensive experiments show that RDCR achieves superior or comparative results under different noise types and levels against existing methods.

In future work, we may incorporate consistency regularization in the space of rotation inputs to eliminate the possible noise in rotation labels. 
In addition, our success hints the potential in applying other Self-SL methods. We can include more auxiliary tasks and labels~\cite{doersch2017multi} to reduce the reliance on observed labels.


\medskip

\small

\bibliographystyle{aaai} 
\bibliography{bibfile1}

\end{document}